\documentclass[10pt,twocolumn]{article}
\usepackage{moreverb,url}

\usepackage[nolist, nohyperlinks, printonlyused, withpage]{acronym}
\usepackage{amsmath}
\usepackage{amssymb}
\usepackage{booktabs}
\usepackage{caption}
\usepackage{graphicx}  
\usepackage{hyperref}
\usepackage{multirow}
\usepackage{placeins}  
\usepackage{subcaption}
\usepackage{wrapfig}
\usepackage{xcolor}

\graphicspath{{img/}}  

\usepackage{algorithm}
\usepackage{algpseudocode}

\makeatletter
\newcommand\fs@betterruled{%
  \def\@fs@cfont{\bfseries}\let\@fs@capt\floatc@ruled
  \def\@fs@pre{\vspace*{7pt}\hrule height.8pt depth0pt \kern2pt}%
  \def\@fs@post{\kern2pt\hrule\relax}%
  \def\@fs@mid{\kern2pt\hrule\kern2pt}%
  \let\@fs@iftopcapt\iftrue}
\restylefloat{algorithm}
\makeatother

\newcommand*\Let[2]{\State #1 $\gets$ #2}  

\usepackage{enumitem}
\newlist{hypotheses}{enumerate}{1} 
\setlist[hypotheses,1]{
    label=\textbf{H\arabic*},
    ref=H\arabic*, 
    itemsep=0pt, 
    parsep=3pt  
}

\newlist{contributions}{enumerate}{1} 
\setlist[contributions,1]{
    label=\textbf{C\arabic*},
    ref=C\arabic*, 
    itemsep=0pt, 
    parsep=3pt  
}

\newcommand{\xhdr}[1]{\vspace{1.7mm}\noindent{{\bf #1.}}}

\newcommand{\xhdrNoPeriod}[1]{\vspace{1.7mm}\noindent{{\bf #1}}}

\usepackage{amsmath,amsfonts,bm}

\def\Figref#1{Figure~\ref{#1}}

\def\eqref#1{equation~\ref{#1}}

\def\1{\bm{1}}

\def\va{{\bm{a}}}

\def\vd{{\bm{d}}}

\def\vp{{\bm{p}}}
\def\vq{{\bm{q}}}

\def\vz{{\bm{z}}}

\def\mQ{{\bm{Q}}}
\def\mR{{\bm{R}}}

\DeclareMathAlphabet{\mathsfit}{\encodingdefault}{\sfdefault}{m}{sl}
\SetMathAlphabet{\mathsfit}{bold}{\encodingdefault}{\sfdefault}{bx}{n}

\newcommand{\R}{\mathbb{R}}

\DeclareMathOperator*{\argmin}{arg\,min}

\DeclareMathOperator{\Tr}{Tr}

\newcommand{\axx}[0]{\text{x}}

\newcommand{\axz}[0]{\text{z}}
\newcommand{\norm}[1]{\left\lVert#1\right\rVert}

\def\valpha{{\bm{\alpha}}}

\def\params{{\bm\theta}}
\def\obstacle{{o}}
\def\design_net{{\mathcal{D}}}
\def\ik_net{{\mathcal{K}}}

\DeclareMathOperator*{\atantwo}{atan2}
\DeclareMathOperator*{\signeddistance}{sd}

\newcommand{\RNum}[1]{\uppercase\expandafter{\romannumeral #1\relax}}

\usepackage{tikz}
\newcommand{\circled}[2][0.75ex]{%
  \tikz[baseline=(char.base)]{
    \node[shape=circle,draw,inner sep=#1] (char) {#2};}}
\newcommand{\smallcircled}[2][0.5ex]{%
  \tikz[baseline=(char.base)]{
    \node[shape=circle,draw,inner sep=#1] (char) {#2};}}
\newcommand{\verysmallcircled}[2][0.2ex]{%
  \tikz[baseline=(char.base)]{
    \node[shape=circle,draw,fill=white,inner sep=#1] (char) {#2};}}
\newcommand{\verysmallcircledcomment}[2][0.12ex]{%
  \tikz[baseline=(char.base)]{
    \node[shape=circle,draw,inner sep=#1] (char) {#2};}}
    
\usepackage{printlen}  

\usepackage[utf8]{inputenc}
\usepackage[T1]{fontenc}
\usepackage{natbib}
\usepackage{authblk}  
\usepackage{hyperref} 
\usepackage{newtxtext, newtxmath}

\renewenvironment{abstract}{%
  \begin{center}%
    \bfseries Abstract
  \end{center}%
  \vspace{-0.5em}%
  \begin{quote}%
}{%
  \end{quote}%
  \vspace{1em}%
}

\usepackage{authblk}
\usepackage[a4paper,margin=0pt]{geometry} 
\usepackage{dblfloatfix}  
\begin{document}

\title{A Design Co-Pilot for Task-Tailored Manipulators}

\author[1,2]{Jonathan K\"ulz\thanks{Corresponding author: \href{mailto:jonathan.kuelz@tum.de}{jonathan.kuelz@tum.de}}}
\author[3]{Sehoon Ha}
\author[1,2]{Matthias Althoff}

\affil[1]{Technical University of Munich, Department of Computer Engineering}
\affil[2]{Munich Center for Machine Learning (MCML)}
\affil[3]{Georgia Institute of Technology}

\date{}


\twocolumn[
  \begin{center}
    \maketitle
  \end{center}
  \begin{abstract}
Although robotic manipulators are used in an ever-growing range of applications, robot manufacturers typically follow a ``one-fits-all'' philosophy, employing identical manipulators in various settings.
This often leads to suboptimal performance, as general-purpose designs fail to exploit particularities of tasks.
The development of custom, task-tailored robots is hindered by long, cost-intensive development cycles and the high cost of customized hardware.
Recently, various computational design methods have been devised to overcome the bottleneck of human engineering.
In addition, a surge of modular robots allows quick and economical adaptation to changing industrial settings.
This work proposes an approach to automatically designing and optimizing robot morphologies tailored to a specific environment.
To this end, we learn the inverse kinematics for a wide range of different manipulators.
A fully differentiable framework realizes gradient-based fine-tuning of designed robots and inverse kinematics solutions.
Our generative approach accelerates the generation of specialized designs from hours with optimization-based methods to seconds, serving as a design co-pilot that enables instant adaptation and effective human-AI collaboration.
Numerical experiments show that our approach finds robots that can navigate cluttered environments, manipulators that perform well across a specified workspace, and can be adapted to different hardware constraints.
Finally, we demonstrate the real-world applicability of our method by setting up a modular robot designed in simulation that successfully moves through an obstacle course.
\end{abstract}
]

\section{Introduction}
\begin{figure}
    \centering
    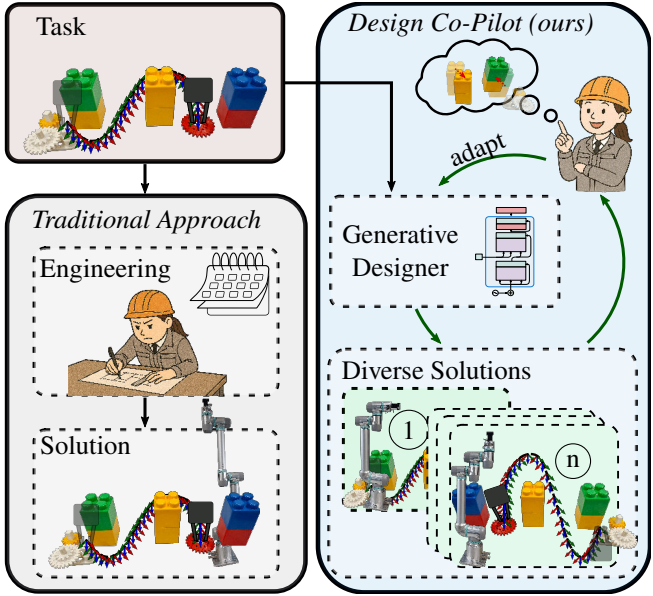
    \caption{%
    Task-tailored manipulators, especially modular robots, are often designed through slow, manual engineering workflows.
    We introduce a generative model that produces diverse task-tailored robots within seconds, enabling rapid exploration of alternatives, paving the way towards supporting engineers in an interactive, co-pilot role.
    }
    \label{fig:title_figure}
\end{figure}
Robotic manipulators are traditionally designed by human experts.
Although engineers can invent and adapt complex machines under strict constraints, such development processes are constrained by long development cycles, the availability of the workforce, and human imagination.
In addition, focusing on standardized solutions allows manufacturers to achieve cost reductions through economies of scale.
As a result, industrially available manipulators are general-purpose machines without being optimized for the specific scenarios in which they are deployed.
Modular reconfigurable robots offer a promising alternative by enabling rapid and cost-effective hardware reconfiguration of standardized components.
However, determining the optimal configuration of pre-manufactured hardware modules for a given task remains a difficult and actively researched problem~\citep{Hoffman2025, Kuelz2025}.
Crucially, the performance of a robot depends both on its morphology and how it can be controlled, making joint optimization of module configuration and robot motion a central challenge.
Traditional approaches rely either on human experience or optimization-based methods that, while systematic, are computationally expensive and hence impractical for rapid design iterations.

Recent advances in generative artificial intelligence (AI) have demonstrated transformative potential for industrial design processes~\citep{Mueller2025}.
Figure~\ref{fig:title_figure} illustrates an emerging design paradigm in which generative models enable rapid exploration and interactive human-AI collaboration, akin to large language models.
As demonstrated by \cite{Schulz2017}, even novice designers can be enabled to create functional robots if given a curated selection of options, highlighting the potential of collaborative human-computer engineering.
Existing generative approaches for robotic design have shown promise in applications such as soft robots~\citep{Wang2023} and individual robot components such as grippers~\citep{Ha2020}.
However, these approaches often struggle to incorporate economic and manufacturability considerations, such as a standardization of parts~\citep{Stella2023}, especially in the absence of application-specific curated datasets.
Moreover, many generative approaches focus on hardware alone, ignoring the challenges that arise in controlling novel designs.
Another line of work focuses on optimizing initial designs through iterative methods~\citep{Ma2021, Xu2021}.
However, optimization-based approaches are computationally expensive and typically converge to single solutions, limiting their utility for rapid design exploration and interactive human-AI collaboration.

Beyond merely generating novel robot morphologies, automated robot design must address a fundamental challenge to enable rapid, interactive design exploration: the intricate coupling between the morphology of a robot and its capacity to execute specific tasks~\citep{Cheney2016}.
The kinematics of a manipulator, particularly the reachability of desired poses in potentially cluttered environments, serves as a decisive criterion for assessing design  quality~\citep{Campos2020,Romiti2023,Hoffman2025}.
However, even minor morphological changes can drastically affect kinematic capabilities, making design evaluation computationally demanding.
Since analytical inverse kinematics solutions are unavailable for most manipulators, their assessment relies on numerical optimization methods that are both slow and potentially incomplete, creating a significant bottleneck in automated design workflows.

This work presents a generative approach that overcomes the aforementioned limitations by jointly training a generative manipulator design network and a corresponding inverse kinematics solver.
Once trained, these networks can rapidly generate specialized manipulators without requiring initial solutions, pre-collected datasets, or lengthy optimization cycles.
Our framework applies to both discrete modular robot configurations and continuous design parameterizations, generating diverse, high-performing manipulator designs in seconds rather than hours or days.
Crucially, these advances enable, for the first time, a true human-AI co-pilot for manipulator design. Designers can interactively explore new manipulator configurations, respond to emerging possibilities, and adjust goals on the fly, ensuring that the design process evolves with the task at hand.

In summary, this work contributes the following aspects to the state-of-the-art in manipulator design:
\begin{itemize}
    \item A generative “co-pilot” framework that maps structured environment representations directly to specialized, manufacturable manipulator designs, thus enabling human-AI design collaboration.
    \item A learning-based approach that leverages differentiable task performance objectives, eliminating the need for pre-collected datasets or lengthy optimization for modular robot composition optimization.
    \item The first deep learning-based framework that jointly optimizes manipulator morphology and inverse kinematics solvers, thus enabling rapid design generation without numerical optimization bottlenecks.
    \item The demonstration of applicability to discrete modular robot configurations and continuous design parameterizations, providing flexibility across different hardware platforms.
    \item A validation showing enhanced task performance and computational efficiency compared to general-purpose manipulators or traditional optimization methods.
\end{itemize}

\section{Related Work}\label{sec:related_work}
In this section, we describe the current methods for co-optimizing the morphology and behavior of robots, generative approaches to hardware design, and previous work on learning-based solvers for inverse kinematics.

\xhdr{Co-Optimization of Morphology and Control}
Inspired by nature, early research by \citet{Sims1994} and \citet{Lipson2000} deployed evolutionary methods to jointly optimize robots and their control.
More recently, \citet{Gupta2021} advanced the concept of evolutionary morphology optimization by learning robot-conditioned control policies via deep reinforcement learning.
\cite{Deroo2023} motivate the co-design of task-specific robots and control with potential energy and cost savings, arguing that general-purpose mechanisms are often overengineered for the scenarios in which they are being deployed.
Complementing these developments, \citet{Wu2025} highlight that design optimization can also lead to better explainability of control failures in motion planning.

In the context of continuously parameterized robots, \citet{Ha2019} and \citet{Schaff2019} leverage reinforcement learning to optimize robot morphologies and their control policies.
However, reinforcement learning methods often suffer from sample inefficiency, as the physics of the underlying systems is not explicitly taken into account and needs to be learned from scratch in simulation.
In contrast, methods based on analytical gradients leverage differentiable system models to replace the trial-and-error strategy applied in black-box optimizers by deploying a fully differentiable framework for optimization.
Being sample-efficient and robust even for large design spaces, they show great potential for the design of physical systems~\citep{Allen2022}.
\citet{Xu2022} have shown that gradient-based control optimization for various fixed morphologies works even over long time horizons.
Especially for soft robots, differentiable models of robots and environments have been successfully applied~\citep{Baecher2021, Matthews2023}, as their inherently freeform design realizes continuous parameterization and optimization.

When designing rigid robots, freeform solutions are either not producible at all or their manufacturing costs exceed the benefit of a specialized robot.
The recent introduction of modular robots with self-programming capabilities by~\cite{Althoff2019} and \cite{Romiti2022} facilitates the rapid realization of novel hardware designs in real-world applications.
For the corresponding discrete module configuration problem, search-based iterative design methods have been employed; for instance, \citet{Liu2020} and \citet{Zhao2020} propose heuristic search algorithms tailored to modular design spaces.
In modular robot design, control-relevant robot properties are often evaluated through kinematic criteria~\citep{Liang2025}, such as the collision-free reachability of desired workspace poses:
\citet{Whitman2020} utilize the value function of a trained reinforcement learning agent as a search heuristic for possible modular robot configurations.
Despite such innovations, evolutionary algorithms remain a standard approach in the optimization of modular robot composition.
\citet{Alattas2019} provide a comprehensive overview of their use in different types of modular robots.
In recent work, \citet{Romiti2023} and \citet{Kuelz2024} have developed customized genetic algorithms for optimizing discrete modular robot morphologies, further extended by \citet{Kuelz2025} to multi-objective optimization.

An alternative approach in rigid robot optimization is to determine a fixed topology and optimize parameters such as specific lengths or radii~\citep{Schaff2019, Ha2019, Hu2022b, Vaish2024}.
~\citet{Ha2017} leverage the implicit function theorem to optimize and refine designs around a local optimum, and ~\citet{Maloisel2023} optimize parameterized character kinematics via dynamic programming.
\citet{Hoffman2025} use mixed integer nonlinear programming and evolutionary strategies to design robots for specific workspaces by varying components continuously in a particular range or by choosing discretely from a catalog.

Although optimization-based methods can produce task-specific solutions, they are computationally expensive, often requiring thousands to millions of iterations for each new task.
This high computational cost prevents engineers from obtaining real-time recommendations during iterative product design.
In contrast, our method enables real-time inference, providing immediate, task-adapted design suggestions.

\xhdr{Generative Hardware Design}
Following the recent success of generative language models, generative design approaches have been proposed in the context of mechanical (robot) hardware.
While tackling challenges similar to traditional co-optimization works, they facilitate rapid prototyping and iterative development by providing multiple, diverse designs for novel tasks quickly.
\citet{Ha2020} employ an autoencoder-based framework for the design of robotic hands.
\citet{Kim2025} propose a generative framework for the multi-objective optimization of jumping robots, based on a fixed initial robot topology.
However, these methods focus solely on robot hardware, assuming that control of the resulting designs is morphology independent or can trivially be inferred.
Many generative approaches have focused on soft and walking robots~\citep{Cheney2014, Wang2023, Wang2023a}, which are naturally compatible with gradient-based optimization due to their free-form design spaces and smooth objective functions.
By comparison, relatively little work has addressed generative design for standardized or modular robotic hardware.
\citet{Hu2022} use generative adversarial networks to synthesize legged robots from modular parts, using evolutionary optimization to generate designs without labeled training data.
In a related study, \citet{Hu2022a} adopt an autoencoder-based approach that relies on model predictive control, introducing a significant computational bottleneck.
Reported computation times in the range of minutes per robot limit the applicability of these approaches to relatively small design spaces.
In contrast, our method is applicable to both continuous and discrete design spaces, with a learned inverse kinematics network enabling the evaluation of thousands of robots per second.

\xhdr{Learning Inverse Kinematics}
The inverse kinematics (IK) problem needs to be solved efficiently to evaluate the capabilities of a manipulator.
Especially in iterative optimization processes, when millions of possible kinematic designs are considered, the runtime of an IK algorithm becomes crucial.
The fastest solution for a single robot is usually provided by analytical methods.
However, methods to derive a closed-form solution for a novel manipulator are only available for robots with specific kinematics, while a general analytic solution algorithm has yet to be developed~\citep{Diankov2010, Elias2025}.

Solvers based on numerical optimization can be applied to arbitrary manipulators, but usually return a single solution only and are slow compared to other approaches.
For these reasons, \citet{Ardizzone2019}, \citet{Ames2022}, and \citet{Li2023b} study learning-based methods for IK for single manipulators.
\citet{Bensadoun2022} propose using Gaussian mixture models to capture the multiplicity of possible solutions to the IK problem.
The work of \citet{Oehsen2020} shows that a simple distal-teaching approach is sufficient to train a neural network that accurately predicts a single IK solution.
Similarly,~\citet{Tenhumberg2023} train a twin-headed network to generate multiple initial solutions for IK in cluttered environments.
While highlighting the potential of learning-based IK solvers, all of the above-mentioned approaches are trained for only a single robot model.
They are therefore not directly applicable to co-optimization procedures.
~\citet{Limoyo2025} introduce a generative graph-based approach that can generalize to various robot kinematics.
While promising, their approach has only been shown to work on a restricted set of kinematics and does not consider self-collisions or collisions with the environment.
In our work, we propose an approach for learning the inverse kinematics of generic manipulators, leveraging a fully differentiable model of their forward kinematics.

\section{Approach}\label{sec:approach}
\begin{figure*}
    \def\svgwidth{\textwidth}  
    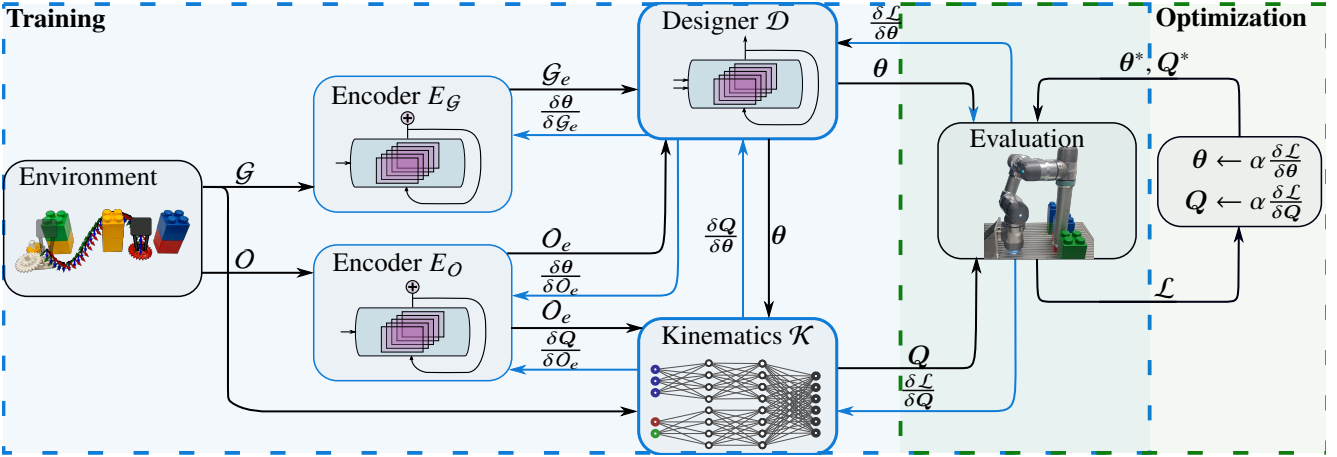
    \caption{
    We train a generative designer network $\design_net$ (Section~\ref{sec:gen_net}) and a kinematics network $\ik_net$ (Section~\ref{sec:ik_net}) on a differentiable loss function (Section~\ref{sec:loss}).
    Environment data is encoded (Section~\ref{sec:encoders}) and fed into $\design_net$ and $\ik_net$.
    To obtain final parameters, we use first-order optimization to find locally optimal parameters $\params^*, \mQ^*$ based on the initial guesses $\params, \mQ$ (Section~\ref{sec:opt}).
    }
    \label{fig:pipeline}
\end{figure*}
Our approach is based on a generative designer network $\design_net$, which takes environment~$\mathcal{O}_e$ and goal~$\mathcal{G}_e$ embeddings as inputs and produces hardware parameters $\params$, tailored to a specific task.
To evaluate robot designs, we leverage a kinematics network $\mathcal{K}$ that produces collision-free IK solutions $\mQ$.
Figure~\ref{fig:pipeline} summarizes the overall architecture.
Section~\ref{sec:problem_statement} introduces the formal problem statement.
Section~\ref{sec:loss} details how we ensure the differentiability of our objective function.
In Section~\ref{sec:encoders}, we explain how we obtain numerical embeddings for environment and goal information.
These are then fed into the designer and kinematics networks introduced in Section~\ref{sec:gen_net} and Section~\ref{sec:ik_net}.
Finally, Section~\ref{sec:opt} describes how the differentiable objective enables fast gradient-based refinement of generated solutions, while Section~\ref{sec:dof_constraint} explains how we constrain robot degrees of freedom and incorporate goal tolerances based on task requirements.

\subsection{Problem Statement}\label{sec:problem_statement}
Given an environment $\mathcal{O} = \{o_1, \dots, o_k\}$ with $k$ obstacles $o_i$ and target end-effector poses \mbox{$\mathcal{G} = \{g_1, \dots, g_m\}$}, we want to find manipulators that can reach all goals without collisions, including self-collisions, at minimal hardware cost.
Formally, we consider a co-optimization problem of the form
\begin{equation}
\begin{aligned}
    \left[ \params^*, \mQ^* \right]
    &= \argmin_{\params, \mQ} \mathcal{L}(\params, \mQ) \text{, where} && \\
    \mathcal{L}(\params, \mQ) &= \sum_{i=1}^m \biggl( \underbrace{w_d d(g_i, \params, \vq_i)}_{\text{distance to goal}} &&+ \underbrace{w_{col} c_{col}(\mathcal{O}, \params, \vq_i)}_{\text{collision cost}} \biggr) \\
    &+ \underbrace{w_h c_h(\params)}_{\text{hardware cost}}    &&+ \underbrace{w_{reg} c_{reg}(\params, \mQ)}_{\text{regularization cost}} \, , \label{eq:loss}
\end{aligned}
\end{equation}
where $\params \in \mathbb{R}^{n \times k}$ parameterize the morphology of a single robot manipulator with $n$ degrees of freedom and \mbox{$\mQ = \begin{pmatrix}
    \vq_1, \dots, \vq_m
\end{pmatrix} \in {R}^{n \times m}$} represents the IK solutions for all $m$ goals.
The distances $d(\cdot)$ between the goals and the end effector, given by the forward kinematics $FK(\params, \vq_i)$, serve as our primary objective.
The collision term guides the optimization toward feasible, collision-free configurations.
We also include a hardware cost term, as well as a regularization term that encourages diversity among the generated solutions.
During training, we evaluate the objective directly on the network outputs $\params$ and $\mQ$.
During evaluation, we treat them as an initial guess and optimize them numerically, leveraging analytical gradients for evaluation.

This work focuses on optimizing robot kinematics without specifying a particular end effector.
Since industrial end effectors constitute static geometric offsets in the forward kinematics, any choice of end effector can be incorporated post-design through a corresponding translation of target poses.
Therefore, we solve the optimization problem in \eqref{eq:loss} with respect to a reference frame attached to the last link of the kinematic chain, with the understanding that end-effector-specific offsets would be applied to the goal poses upon selection of a particular tool.

\subsection{Differentiable Objective}\label{sec:loss}
As we want to optimize both the designer $\design_net$ and the kinematics network $\ik_net$, based on gradients of the loss term $\mathcal{L}$, we need it to be differentiable with respect to parameters $\params$ and joint angles $\mQ$.
We represent goal poses as $g_i = (\vp_i, \mR_i)$, where $\vp_i \in \R^3$ denotes the desired position and $\mR_i \in SO(3)$ represents the target orientation.
The weighted distance function $d(\cdot)$ between two poses is defined as:
\begin{align}
    d(g_g, g_{FK}) &= \underbrace{\norm{\vp_g - \vp_{FK}}}_{\text{Euclidean distance}} \label{eq:pose_distance} \\
    &+ w_r \underbrace{\arccos \left( \frac{ \Tr \left( \mR_g^{-1} \mR_{FK} \right) - 1}{2} \right)}_{\text{rotational distance}} \, . \nonumber
\end{align}
According to Euler's rotation theorem, any rotation $\mR \in SO(3)$ can be represented by an angle $\varphi \in [0, \pi]$ and a unit axis of rotation~\citep{Bauchau2003}.
We compute the rotational distance between the desired and actual end-effector pose as the angle \mbox{$\varphi = \arccos \left( \frac{1}{2} \left(\Tr \left( \mR \right) - 1\right) \right)$} and disregard the rotation axis.
As $\arccos(x)$ has infinite gradients for $x \in \{-1, 1\}$, during training and optimization, we clip the trace of the term between goal and end-effector orientation $\mR_g^{-1} \mR_{FK}$, thus ignoring errors smaller than $0.2 \deg$.

Inspired by optimization-based motion planning techniques as presented by~\citet{Ratliff2009} or \cite{Schulman2014}, we define a collision loss that decreases exponentially with the signed distance $\signeddistance \left( \obstacle_1, \obstacle_2 \right)$ between two collision objects and is zero above some threshold $t$:
\begin{align}\label{eqn:signed_distance_loss}
    \mathcal{L}_{col} = \begin{cases}
        \frac{1}{1 - e^{-1}} \left( e^{- \frac{\signeddistance \left( \obstacle_1, \obstacle_2 \right)}{t} } - e^{-1} \right) &\text{, if } \signeddistance \left( \obstacle_1, \obstacle_2 \right) \leq t \\
        0 &\text{, otherwise.}
    \end{cases}
\end{align}
To efficiently compute collisions, we approximate each obstacle as a set of spheres and each robot link as a set of capsules, which allows analytic evaluation of the signed distance function and the corresponding gradients~\citep{Thumm2022, Sundaralingam2023}.
The same approach is compatible with other signed distance-based obstacle representations, provided that gradients are available~\citep{Zimmermann2022, Millane2024}.
The resulting cost $c_{col}$ is defined as the summed pairwise collision loss over all relevant collision pairs, as detailed in Appendix~\ref{apdx:geometry_approx}.

\subsection{Numerical Embeddings}\label{sec:encoders}
With two encoder networks, we create fixed-dimensional latent representations for the environment $\mathcal{O}$ and goals $\mathcal{G}$.
To embed arbitrary sets of goals and obstacles, we employ the set transformer architecture of~\citet{Lee2019}, which provides permutation-invariant embeddings without discretization or assumptions about set size.
Following~\citet{Joshi2025}, this formulation is functionally equivalent to but computationally more efficient than graph neural network encoders that have been explored in related contexts~\citep{Khan2020, Lai2025}.

We construct separate embeddings for obstacles and goals.
Obstacle embeddings $\mathcal{O}_e$ are shared between the designer and the kinematics network, while goal embeddings $\mathcal{G}_e$ are provided only to the designer.
The kinematics network instead processes each goal individually when solving the corresponding inverse kinematics problem.
Unlike approaches as presented by \citet{Hu2022a} or \citet{Hu2022}, which condition morphology and control on a single environment embedding, our method supplies the designer with aggregated goal information, while the kinematics network handles goals separately.

To support the learning of semantically meaningful embeddings early in the training, we add a reconstruction loss term to the overall training objective.
Ideally, numerical embeddings allow us to reconstruct whether a certain point (or pose) is close to a goal or obstacle, respectively.
Let $\mathcal{I}$ be the set of all goal indices $i$ of goals $g_i$.
We define the closest-point distances to sets of goal positions, goal orientations, and obstacles as
\begin{align}
    d_{pos}(\vp_q) &= \min_{i \in \mathcal{I}} \norm{\vp_i - \vp_q} \\
    d_{ori}(\mR_q) &= \min_{i \in \mathcal{I}} \arccos \left( \frac{ \Tr \left( \mR_i^{-1} \mR_{q} \right) - 1}{2} \right) \\
    d_{obs}(\vp_q, \mathcal{O}) &= \min_{o \in \mathcal{O}} sd(\vp_q, o) \, ,
\end{align}
where $sd(\vp_q, o)$ is the signed distance between a query point $\vp_q$ and obstacle $o$.
A successful reconstruction of these distances indicates that the embeddings effectively allow for partitioning the environment into relevant and irrelevant spaces, as well as occupied and free spaces.
We train three separate, simple feedforward networks jointly with the encoders to predict the point-set distances for randomly chosen query points and orientations and compute the reconstruction loss as the mean squared error between ground truth and prediction.

\subsection{Designer Network}\label{sec:gen_net}
\begin{figure}
    \def\svgwidth{0.75\linewidth}
    \centering
    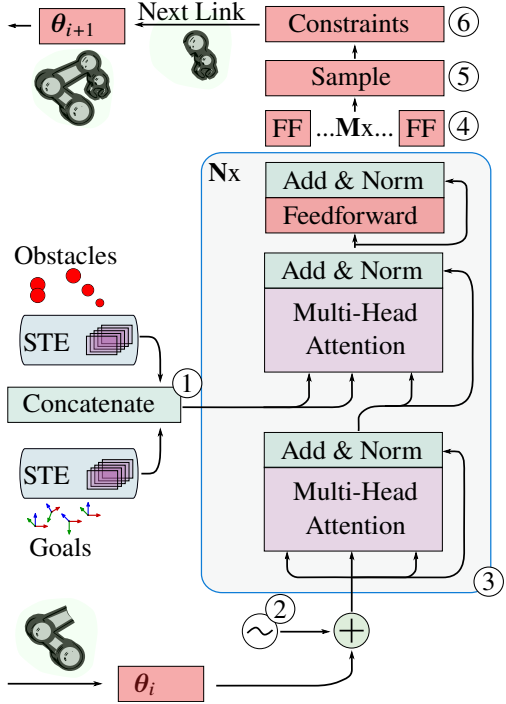
    \caption{
    The model architecture of the designer network $\mathcal{D}$.
    A robot is generated as a sequence of links, connected by revolute joints.
    We generate environment embeddings leveraging set transformer encoders (STE).
    To obtain the next link of an unfinished robot, we encode the existing kinematic structure and combine it with goal and obstacle embeddings using standard transformer blocks as introduced by \cite{Vaswani2017}.
    }
    \label{fig:designer_architecture}
\end{figure}
\begin{algorithm}[t!]
\caption{Generate a Robot}
\label{alg:gen_robot}
\begin{algorithmic}[1]
    \Function{new\_robot}{$\mathcal{G}_e, \mathcal{O}_e, dof$, mode}
    \Let{seed}{sample uniformly at random}
    \Let{$\params_0$}{initialize parameters from seed}
    \Let{$enc$}{concatenate($\mathcal{G}_e, \mathcal{O}_e$)} \Comment{\verysmallcircledcomment{1}}
    \For{$i = 1 \dots dof$}
        \Let{$x$}{positional\_encoding($\params_{i-1}$)} \Comment{\verysmallcircledcomment{2}}
        \Let{$y$}{decoder($x, enc$)} \Comment{\verysmallcircledcomment{3}}
        \Let{$\bm{\varphi}$}{$FF(y_i)$} \Comment{\verysmallcircledcomment{4}}
        \Let{$\hat{\params}_{raw}$}{sample from $p_{\text{mode}}(\hat{\params} \mid \bm{\varphi})$} \Comment{\verysmallcircledcomment{5}}
        \Let{$\hat{\params}$}{apply constraints $(\hat{\params}_{raw})$} \Comment{\verysmallcircledcomment{6}}
        \Let{$\params_i$}{stack$\left(\params_{i-1}, \hat{\params}\right)$}
    \EndFor
    \Let{$\params$}{$\params_\text{dof}.pop(0)$} \Comment{remove initial seed}
    \State \Return{$\params$}
    \EndFunction
\end{algorithmic}
\end{algorithm}

The designer network $\design_net$ generates a robot $\params$ that can reach all goals $\mathcal{G}$ without colliding with itself or obstacles $\mathcal{O}$.
We formulate this as a sequential decision-making problem, where each added link must account for prior assembly steps and the overall design context.
Prior generative models for discrete hardware, such as presented by \citet{Etesam2025} and \citet{Xu2022a}, are typically trained on large datasets of ground-truth examples.
Such data does not exist for the task-tailored design of robotic manipulators.
Instead, as illustrated in Figure~\ref{fig:pipeline}, we directly train the designer network on gradient information obtained through the differentiable loss defined in~\eqref{eq:loss}.

Our model builds on the transformer architecture by~\citet{Vaswani2017}, which naturally captures the mutual dependencies inherent in sequential design processes.
Figure~\ref{fig:designer_architecture} illustrates the overall model architecture, while Algorithm~\ref{alg:gen_robot} details the sequential generation process for a single robot.
Starting from a randomly sampled seed, the network iteratively adds links to the parameter representation until the specified degrees of freedom are reached.
At each step, concatenated environment and goal embeddings condition the designer decoder, which predicts distribution parameters $\bm{\varphi}$.
Here, $\bm{\varphi}$ refers to the parameters that define the probability distribution from which the next link is sampled, e.g., logits for a categorical distribution, or the mean and variance for a normal distribution.
The hardware parameters are sampled from \mbox{$p(\params \mid \bm{\varphi})$} and immediately processed with differentiable design constraints, ensuring physically valid assemblies.
The exact form of $p$ depends on the chosen parameterization, which we describe in the following paragraph.

\begin{figure*}
    \centering
    \def\svgwidth{\textwidth}  
    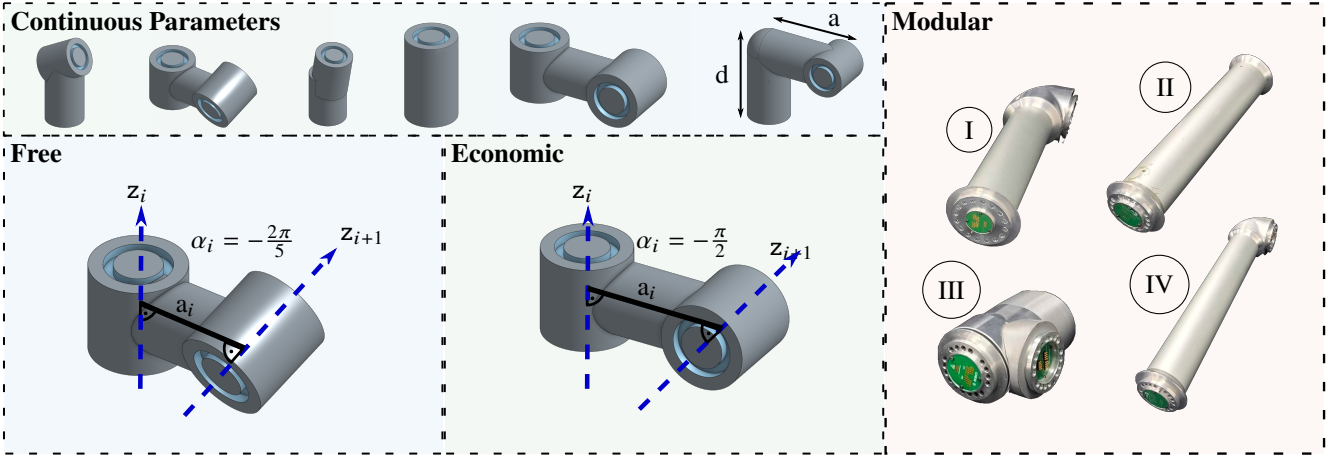
    \caption{
    The free and economic design mode can be used with any parameterization that allows a continuous geometry interpolation, resulting in a broad range of realizable links (top left).
    In this work, we parameterize link lengths along and orthogonal to the previous joint axis with parameters $a$ and $d$; the angle between consecutive joint axes is parameterized by $\alpha$.
    In contrast to the free mode (left), the economic mode enforces hybrid constraints, such as orthogonal consecutive joint axes, during evaluation and at the final training step (center).
    The modular mode is limited to a discrete set of module types (right).
    }
    \label{fig:design_modes}
\end{figure*}

The deployment of task-tailored manipulators in real-world tasks requires economic production.
To this end, we impose constraints on the generated designs that determine how different parameterizations are mapped to robot hardware.
We distinguish three different design modes, accounting for varying levels of versatility and standardization:

\xhdrNoPeriod{The free design mode} allows the use of any continuous geometric parameterization for each robot link.
An example is shown in \Figref{fig:design_modes}, where individual geometries are defined by two lengths $a$ and $d$, and a relative rotation between consecutive joint axes around a common normal $\alpha$.
Each free parameter is sampled from a normal distribution, with mean and standard deviation predicted by the designer network.
To enforce parameter bounds, the samples are passed through a logistic function.
Any monotonic squashing function could be used in principle, but in this work, we use the standard Sigmoid function.
Robots generated in free design mode can, in principle, require fully custom hardware.
While such designs may not be economical for short-term use, they are valuable in domains where task-specific performance is critical, and they provide insight into the limits of unconstrained morphologies.

\xhdrNoPeriod{The economic design mode} describes hybrid continuous-discrete design parameterizations.
Design parameters can often take either the value zero, indicating that a specific piece of hardware is unused, or lie within one or multiple defined intervals.
For example, the geometry offset $a$ in Figure~\ref{fig:design_modes} realizes the economic manufacturing of robot links only within the hybrid constraint $a \in \left( \{0\} \cup [-l_{\max}, -l_{\min}] \cup [l_{\min}, l_{\max}] \right)$.
Here, the absolute value of $a$ determines the size of the geometry, while the sign indicates its orientation.
In addition, discrete constraints, such as requiring consecutive joint axes to be orthogonal or parallel, must often be maintained.
While continuous parameters can be constrained to specific intervals as in free design mode, discrete choices require a differentiable mechanism.
To this end, we use the Gumbel-Softmax trick~\citep{Jang2017}, which enables the designer network to differentiably sample discrete classes during training.
In economic design mode, the network produces soft class probabilities for discrete choices during training.
For training, we compute a continuous approximation of the design parameter by taking a weighted sum of the class values using these probabilities.
A temperature schedule ensures that the network converges to valid discrete selections at the end of training.
For evaluation, discrete classes are sampled at all times.

\xhdrNoPeriod{The modular robot mode} is the most constrained design mode we propose.
In this setting, the designer network assembles pre-manufactured robot modules, restricting the design space to fully discrete choices.
Discrete module selections are sampled using the Gumbel-Softmax trick; since interpolation between modules is not meaningful, we employ a straight-through estimator \citep{Jang2017, Bengio2013} for both training and evaluation to allow gradient-based optimization.
While limiting the design space the most, the modular robot mode allows one to assemble the designed robot from the available hardware within minutes of algorithm termination.

\xhdrNoPeriod{For all modes}, we aim for a generative design process that allows human engineers to explore a variety of optimized solutions for a specific problem setting.
To this end, we introduce several strategies to promote design diversity and prevent premature convergence.
During training, we generate $r$ distinct robot designs per environment and compute a loss based on their similarity.
Let $\bm{\theta}_{i, e}$ denote the $i$-th robot design parameters generated for an environment $e$, and let $b$ be the mini-batch size used for training.
The robot similarity loss is computed as
\begin{align}
    \mathcal{L}_{sim} = & \frac{2 w_g}{b r (r-1)} \sum_{e=1}^b \sum_{i=1}^{r-1} \sum_{j=i+1}^{r} \norm{\bm{\theta}_{i, e} - \bm{\theta}_{j, e}}_1 \\
    + & \frac{2 w_e}{b (b-1) r^2} \sum_{e=1}^{b-1} \sum_{f=e+1}^{b} \sum_{i=1}^{r} \sum_{j=1}^{r} \norm{\bm{\theta}_{i, e} - \bm{\theta}_{j, f}}_1 \nonumber \, .
\end{align}
The first term penalizes the similarity among robots generated for the same environment, encouraging the exploration of distinct high-performing designs for a single task context.
The second term prevents the model from collapsing to a small set of generic designs, encouraging task-specific specialization.
We use the L1-norm as a similarity measure because it yields unit-norm gradients even when parameters are close, which helps to recover from early mode collapse.
During training, the similarity loss is included in the total loss as part of the regularization term $c_r$.

To enable the network to generate diverse parameters, we introduce multiple sources of randomness.
First, we train $M$ independently initialized feedforward networks that map parameter encodings to distribution parameters.
Each network generates the parameters for $r / M$ robots.
Second, randomness is introduced by sampling from a normal distribution for continuous parameters and a Gumbel distribution for discrete ones, as detailed in Section~\ref{sec:gen_net}.
Third, rather than using a fixed start-of-sequence token, we initialize the parameter vector $\bm{\theta}$ with uniform random noise prior to generating the first link.
During evaluation, we keep all sources of randomness active to generate a diverse set of solutions, enabling richer exploration and selection of designs post-training.

\subsection{Kinematics Network}\label{sec:ik_net}
The kinematics network $\ik_net$ predicts inverse kinematics solutions ${q_1, \dots, q_n}$ for goals ${g_1, \dots, g_n}$, an obstacle embedding $\mathcal{O}$, and robot parameters $\params$.
We follow~\cite{Tenhumberg2023} and use a multi-headed feedforward network with $N$ heads to capture the multi-modality of inverse kinematics solutions.
To encourage diverse outputs and avoid mode collapse, we add a weighted cosine similarity term to the cost $c_r$:
\begin{align}
c_{r, IK} &= \frac{2 w_{IK}}{\left| \mathcal{I} \right| N (N - 1)} \sum_{i \in \mathcal{I}} \sum_{h=1}^{N-1} \sum_{k=h+1}^{N} \frac{\vq^\top_{i, h} \, \vq_{i, k}}{
\norm{\vq_{i, h}} \norm{\vq_{i, k}}}.
\end{align}
This ensures that each head produces distinct solutions, providing a simple mechanism to represent multiple valid IK solutions per goal.
During training, the loss is computed across all heads to encourage exploration of multiple solution modes; during evaluation, we use the head with the lowest loss to obtain the final IK solution.
Rather than directly predicting joint angles, the network returns two values $(x_i, y_i)$ for each joint.
We compute the corresponding joint angle as $q_i = \atantwo(x_i, y_i)$, which ensures a continuous representation while limiting the joint angles to the interval $[-\pi, \pi]$.

\subsection{Optimization}\label{sec:opt}
The differentiable loss function allows gradient-based refinement of both joint angles and continuous manipulator parameters.
We first take the joint angles predicted by the kinematics network $\ik_net$ and refine the inverse kinematics solutions using a pseudo-inverse of the manipulator’s geometric Jacobian.
Finally, the refined joint angles and continuous design parameters are used to warm-start an Adam optimizer~\citep{Kingma2015} for joint co-optimization.

\subsection{Constraining Robot Degrees of Freedom}\label{sec:dof_constraint}
For all problems, the number of degrees of freedom (dof) is determined a priori.
While this constrains the morphologies considered in a single training, training separate networks for each dof offers several advantages.
First, it allows explicit observation of the trade-off between the number of degrees of freedom and task performance.
Second, it keeps the output of $\ik_net$ and $\design_net$ constant in dimension, simplifying interpretation, debugging, and reproducibility.
Finally, it enables the use of separate networks with different objectives depending on the desired dof.

Some tasks, such as handover, welding, or drilling, do not require specifying all degrees of freedom of a robot, allowing the use of underactuated manipulators.
To accommodate this, we introduce two types of goal tolerances that relax the reachability requirement.
The first, a tolerance for rotational symmetry, allows for arbitrary rotations around the end effector $\mathtt{z}$-axis $\vz$ and is incorporated by replacing the rotational distance in \eqref{eq:pose_distance} with
\begin{align}
\frac{1 - \vz \mR_{g}^{-1}\mR_{FK} \vz^\top}{2} \, .
\end{align}
The second, a position-only tolerance, considers only the Euclidean distance and is achieved by setting $\omega_r = 0$ in \eqref{eq:pose_distance}.

\section{Experiment Setup}\label{sec:setup}
In Sections~\ref{sec:parameterization} and~\ref{sec:hardware_modules}, we introduce the parameterization used to assess our approach.
Sections~\ref{sec:pretrain} and \ref{sec:training} describe training procedures.
Finally, Section~\ref{sec:experiment_setup} describes the research hypotheses that we analyze in our experiments.
The source code for all experiments will be published upon acceptance of this article.

\subsection{Morphology Parameterization}\label{sec:parameterization}
For free and hybrid design modes, we parameterize the morphology of a manipulator with parameters $\params$ inspired by Denavit-Hartenberg parameters~\citep{Denavit1955, Robotics2009Kinematics}.
A robot with $n$ degrees of freedom is parameterized by the $n \times 3$ matrix $\params = \begin{pmatrix} \vd & \va & \valpha \end{pmatrix}$.
As realizing the designed morphologies from available, cost-effective materials should be possible, we enforce a piecewise cylindrical manipulator morphology.
To construct the geometry of the $i-\text{th}$ link of a robot, we start from the reference frame $i-1$ and attach a cylinder of fixed radius $r$ and length $d_i$ in direction $\axz_{i-1}$, followed by a cylinder of length $a_i$ in direction $\axx_i$.
If $a=0$ or $d=0$, the parameter is ignored.
The joints are oriented according to the Denavit-Hartenberg convention.
Further, for the economic design mode, we enforce \mbox{$d_i \in \{0\} \cup [0.1, 0.4]$}, \mbox{$a_i \in \{0\} \cup [-0.4, -0.1] \cup [0.1, 0.4]$}, and $\alpha_i \in \{0, \frac{\pi}{2}, -\frac{\pi}{2} \}$.
Examples of the resulting geometries are shown in Figure~\ref{fig:design_modes} and Appendix~\ref{apdx:dh_to_links}.

\subsection{Hardware Modules}\label{sec:hardware_modules}
We evaluate the modular design mode using industrially available robot modules.
They consist of joints and a long and short static extension in two different geometries, each of which realizes the mounting of consecutive joints such that their axes are orthogonal or parallel.
Figures~\ref{fig:example_scene} and~\ref{fig:robco_duplo} show robots assembled from the available modules.
Two joints can be connected directly or using exactly one static extension.
The overall design space for a six-degree-of-freedom robot thus consists of $5^6 = 15625$ different robots that can be constructed from the available modules.
For the hardware setup presented in Section~\ref{res:realworld}, we had access to two static extensions of each type.

\subsection{Pretraining}\label{sec:pretrain}
Before starting the training procedure, we pretrain the kinematics network $\ik_net$.
We randomly generate a robot parameterization $\params_{r}$ and joint angles $\vq_r \in \mathbb{R}^n$ for each step.
Using these data, goals $\mathcal{G}$ can be easily created by computing forward kinematics $FK \left(\params_{r}, \vq_{r} \right)$ and used for training of $\ik_net$.
By ensuring that the randomly sampled parameters follow the constraints imposed on the designer, we train the kinematics network on samples of the whole set of possible kinematics that can occur during the training phase.

\subsection{Training}\label{sec:training}
We sample random parameters $\params_r$ and joint angles $\mQ_r$ to construct goals $\mathcal{G}$ by computing the forward kinematics of the random parameters.
We randomly sample $n_o$ spherical obstacles in the robot workspace and discard all that intersect with any goals.
While our approach supports diverse geometric primitives for obstacle representation, we restrict the experiments to spheres and capsules in evaluation.
We obtain the parameters $\params$ and $\mQ$ from the forward pass of the designer and kinematics network, respectively.
Both networks are updated using the loss introduced in Section~\ref{sec:approach}.
To avoid overfitting $\ik_net$ to the outputs of $\design_net$, we reduce its learning rate and continue to use randomly generated parameters $\params_r$ for a fraction of the training steps.
Details about these and further hyperparameters, as well as compute resources, and training times are listed in Appendix~\ref{apdx:training_details}.

\subsection{Experiments}\label{sec:experiment_setup}
In Sections~\ref{res:vs_baselines}-\ref{sec:design_constraints}, we perform three numerical experiments in different settings to analyze the following hypotheses:
\begin{hypotheses}
  \item Our approach efficiently generates high-quality robot designs for cluttered environments, outperforming general-purpose manipulators. \label{hyp:1}
  \item The designer network adapts to different task requirements and can generate robots to work in specified task spaces. \label{hyp:2}
  \item Our approach works under different manufacturability constraints. \label{hyp:3}
\end{hypotheses}
In Section~\ref{sec:ablation_encoder}, we perform an ablation study to validate the effectiveness of the set transformer architecture for environment encodings.
Finally, we demonstrate how our results can be transferred to real-world settings in Section~\ref {res:realworld}, where we optimize a modular manipulator to follow a collision-free path through an obstacle course.

\section{Results}\label{sec:result}
For each of the following experiments, we generate eight candidate robots with four initial solutions for each inverse kinematics problem and optimize them in parallel.
We report the runtimes for the complete optimization and results for the best robot and joint angles, respectively.
Unless stated otherwise, all results correspond to robots that are neither in self-collision nor in collision with the environment.
Unlike some approaches that jointly optimize robot base placement alongside the manipulator~\citep{Romiti2023, Mayer2025}, this work focuses solely on hardware generation.

 \subsection{Manipulators for Cluttered Environments}\label{res:vs_baselines}
 \begin{figure}
     \centering
     \includegraphics[width=\linewidth]{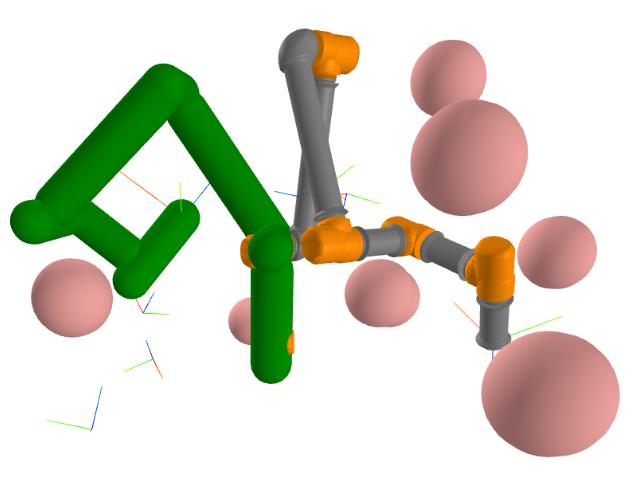}
     \caption{
     We randomly sample eight obstacles (red) and eight goals.
     To evaluate a robot on the problem, we numerically compute a collision-free inverse kinematics solution.
     The figure shows a modular robot and its collision geometry approximation (green) for two different joint configurations.
     }
     \label{fig:example_scene}
 \end{figure}
\begin{figure*}
    \centering
    \includegraphics[width=0.9\linewidth]{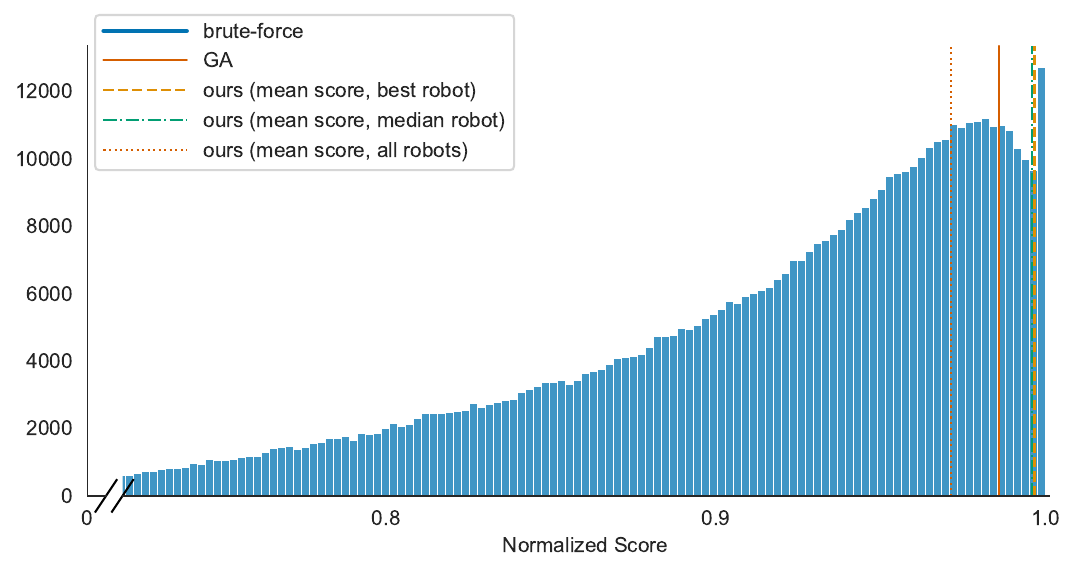}
    \caption{
We compare our approach to a genetic algorithm and a brute-force method by aggregating and normalizing performance across $36$ randomly generated cluttered environments. In every task, we identify at least one robot among the top $10\%$, and consistently find multiple robots that outperform the genetic algorithm baseline.
}
    \label{fig:fitness_score_distributions}
\end{figure*}
We set up $36$ randomized tasks by sampling eight goal poses and eight obstacles, each within a maximum distance from the robot base.
Figure~\ref{fig:example_scene} shows one of the tasks together with a robot design generated and the overapproximation of its collision geometry.
    Interactive visualizations for all tasks are available at \href{https://redirect.cps.cit.tum.de/rd-kuelz}{redirect.cps.cit.tum.de/rd-kuelz}.
We compare robots generated with our approach against near-optimal brute-force solutions.
While brute force is infeasible for large design spaces or time-critical applications, the restricted number of modules in our experiments allows exhaustive evaluation, providing a strong baseline and upper bound for heuristic and evolutionary methods.
To support this, we also compare against a genetic algorithm run separately for each task.

To obtain the brute-force baseline, we evaluate each six-degree-of-freedom robot that can be assembled from the modules introduced in Section~\ref{sec:hardware_modules}.
Each robot and goal is tested with four initial joint configurations, followed by the same numerical inverse kinematics solver and optimization used in our method.
Although randomized IK initialization does not allow us to make hard claims about the optimality of the found solutions, our experience has shown that four initial guesses are usually sufficient to detect a collision-free IK solution if it exists.
For better comparability between tasks, throughout this section, we compute a normalized performance score based on the task loss, where $0$ corresponds to the worst and $1$ corresponds to the best solution found with the brute-force approach.
For evaluation, we consider a goal as solved if the pose error between the end-effector pose and the desired pose is below $1$mm and $1\deg$.
A task is considered solved if all goals can be reached with the corresponding accuracy.
In this section, we aggregate results over all tasks considered; Appendix~\ref{apdx:bf_details} provides an overview of individual task performances.

On average, generating eight candidate robots for a single task with the designer network takes $288$ milliseconds, followed by $9.31$ seconds to optimize the four initial IK guesses for all eight robots.
In the remainder of this section, we only report results for the best of all eight solutions, if not explicitly stated otherwise.
By evaluating $128$ robots in parallel on a GPU, the brute-force approach requires an average of $71$ minutes per task.
The genetic algorithm requires an average of $59$ minutes to run $100$ generations with a population of $128$ for each task.
Each candidate solution is evaluated by computing the distance and collision loss terms introduced in~\eqref{eq:loss}.

Figure~\ref{fig:fitness_score_distributions} shows the distribution of aggregated scores in all $36$ tasks for the brute-force approach, alongside the mean scores for solutions obtained with a genetic algorithm and with our approach.
Most realizable modular robots fail to reach one or multiple goals in the tasks without colliding, producing a long-tailed score distribution.
Among all possible $6$-dof robots, $14.83\%$ achieve a score of at least $0.98$, and on average $99.8$ of the $15625$ robot designs ($0.64\%$) achieve a score of $0.999$ or better per environment, which we consider optimal.
Due to the randomized problem generation, even the brute-force approach was unable to find valid solutions in some cases; overall, it solved $97.57\%$ of goals and $30$ out of the $36$ tasks.

Taking into account the eight robots generated per task, our approach achieves an average best score of $0.997$, an average median score of $0.996$, and a mean score of $0.971$ between the eight robots per task.
In $22$ of the $36$ tasks, the designer network finds a robot out of the set of solutions considered optimal, including $4$ tasks that were solved by neither approach.
Relative to the brute-force distribution, the best solution of our approach lies in the $97.0^\text{th}$ percentile on average, with the lowest being $93.4^\text{th}$.
This corresponds to a mean accuracy of $5.2$mm and $0.13 \deg$, aggregated over all $288$ goals.
The relatively high average position error is largely due to a subset of goals that are obstructed by obstacles and thus physically unreachable for a particular robot.
Of all $288$ goals, $249$ $(86.6\%)$ could be solved with our approach, i.e., we identified a robot and joint angles to reach them within $1$mm and $1 \deg$.
In practice, a human engineer could slightly adjust task requirements to increase coverage: an additional $12$ tasks (bringing the total to $83\%$) could be solved if one or two goals were shifted by at most $10$cm.

In contrast, genetic optimization yields an average score of $0.986$, corresponding to only 7 tasks where an optimal robot is found.
Relative to the brute-force distribution, the genetic algorithm converges towards a solution in the $89.4^\text{th}$ percentile on average, with the lowest being $45.9^\text{th}$.
This corresponds to $195$ goals that could be solved across tasks.

Next, we compare the designs generated with our approach to commonly used industrial manipulators in the same environments.
Specifically, we reconstruct the kinematics of the UR15, UR20, and ABB IRB 2600 robots based on their publicly available Denavit–Hartenberg parameterizations.
These robots were selected because they are similar in size to the manipulators generated by our method.
To ensure a fair comparison, the joint limits of the industrial robots are ignored. 

Table~\ref{tab:dh_comparison} shows the performance of the industrial manipulators across all tasks.
None outperformed the best modular manipulator on any task.
The numerical optimization of the inverse kinematics resulted in average orientation errors below $1\deg$ for all robots; however, the average position errors of the industrial manipulators are consistently higher than for the robots generated with our approach.
In particular, their geometries prevented them from reaching certain goal poses without colliding with obstacles, leading to poor performance in some scenarios, especially for the ABB2600 robot.
This is reflected in the number of goals and tasks solved: the generated robots perfectly reached all goals in 50\% of tasks, while the best industrial manipulator, the UR15, only solved 13 of $36$ tasks.
The performance gap is partly moderated by the constraints of our modules: for instance, creating a compact wrist like that of the UR robots is not possible without making the overall robot too small, since the longest links for both UR robots exceed the size of our biggest modules.

\begin{table}
    \small\sf\centering
    \caption{Comparison between designs generated with our approach and three industrial manipulators.\label{tab:dh_comparison}}
    \begin{tabular}{rccc}
    \toprule
         &  Solved Tasks & Solved Goals & Error ($\varnothing$) \\
    \midrule
    Ours &    $\bm{50}\%$ & $\bm{86.56}\%$ & $\phantom{0}\bm{5.20}$mm \\
    UR15 &    $36\%     $ & $    82.29\% $ & $     \phantom{0}7.27$mm \\
    UR20 &    $33\%     $ & $    77.43\% $ & $     \phantom{0}7.62$mm \\
    ABB2600 & $28\%     $ & $    75.70\% $ & $               20.50$mm \\
    \bottomrule
    \end{tabular}
\end{table}

The results demonstrate that our approach generates high-quality designs for a given task.
Although reliance on numerical methods forbids claims of optimality, the brute-force baseline provides a strong empirical reference.
It yields solutions that are, to the best of our knowledge, the closest achievable given current methods.
Although our results indicate that, in general, multiple different robots are feasible to navigate a specified environment, they also highlight that no single manipulator performs optimally across all scenarios, confirming our approach to task-tailored robot design.
The runtime of exhaustive or population-based optimization methods often increases with the size of the design space, which grows exponentially with the number of available modules.
In contrast, our method offers constant-time inference, making it particularly suitable for scenarios where a one-time training phase is acceptable but fast, repeated inference is required.
In this experiment, generating and optimizing robots with our approach achieved speedups of several hundred times compared to the baselines, while still reaching performance close to the upper bound of exhaustive algorithms.
\textbf{This confirms hypothesis~\ref{hyp:1}.}

 
 \subsection{Adapting to Task Requirements}\label{res:goal_tolerances}
 \begin{figure}
    \centering
    \includegraphics[width=0.8\linewidth]{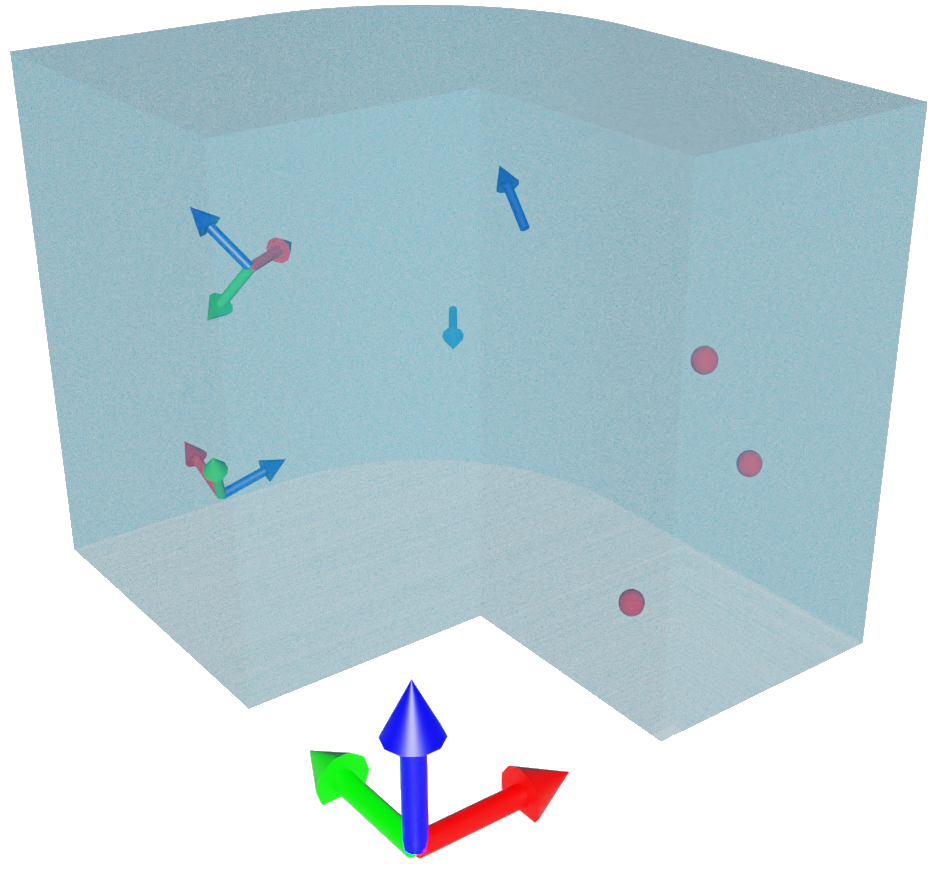}
    \caption{
    The shaded area shows the given workspace $\mathcal{W}$ with maximal extension of $0.8m$, $1.0m$, and $0.8m$ along the $\texttt{x}$, $\texttt{y}$, and $\texttt{z}$-axes, respectively.
    Within $\mathcal{W}$, we sample three kinds of goals:
    Full poses in $SE(3)$ (left) require six degrees of freedom to be reachable in the general case.
    We create five-dimensional goals (center) by adding a rotational symmetry tolerance.
    Position-only goals are specified by coordinates in $\R^3$ (right).
    }
    \label{fig:our_workspace}
\end{figure}
In this section, we analyze how the designer adapts to different goals.
 We adopt an experiment proposed by \citet{Hoffman2025}, in which a robot is synthesized to cover a complete workspace $\mathcal{W}$.
 Figure~\ref{fig:our_workspace} provides an overview of the workspace and the different goal modes (pose, rotational symmetry, and position only) we evaluate.
In Appendix~\ref{apdx:experiment_comparison}, we compare our setup and the one presented by \citet{Hoffman2025}.

\xhdr{Adapting to Goal Tolerances}
 We randomly sample $1000$ positions $\vp_i \in \mathcal{W}$ with random orientations $\mR_i \in SO(3)$ each.
We compare six network variants in modular mode, combining two robot configurations (five- and six-DOF) with three goal types (pose, rotational symmetry, and position-only).
These goal types correspond to the tolerances introduced in Section~\ref{sec:design_constraints}, motivated by practical relaxations of reachability requirements.

\begin{figure}
    \centering
    \includegraphics[width=\linewidth]{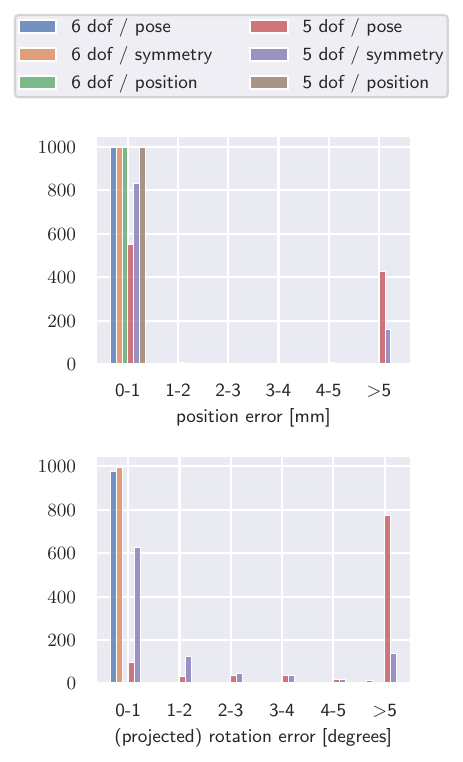}
    \caption{
    We randomly sample 1000 poses within a given workspace $\mathcal{W}$ and design a robot to reach them.
    In \textit{pose} goal mode, the goals are poses in $SE(3)$, in the \textit{symmetry} mode, we allow arbitrary rotations around the end-effector $\mathtt{z}$-axis, and in \textit{position} mode, we optimize for the Euclidean distance between end effector and goal only.
    }
    \label{fig:reachability_results}
\end{figure}

 The generation and optimization of a robot and all 1000 IK solutions take $19.7$ seconds on average.
 Figure~\ref{fig:reachability_results} shows the error distributions in all trials.
 When considering the position only, the generated five-degree-of-freedom and six-degree-of-freedom robots can reach all 1000 poses with an error below $1$mm.
When considering goals with a tolerance for rotational symmetry, we observe a higher overall precision with the six-degree-of-freedom robot that reaches 966 poses with errors below $1$mm and $1\deg$.
Still, even the five-degree-of-freedom robot can reach $55.8\%$ of all goals with a precision better than $1$mm and a projected angular distance of $1\deg$, respectively.
Naturally, a five-degree-of-freedom robot cannot cover a six-dimensional workspace, so we observe high orientation errors for five-dof robots generated for pose goals.
Finally, the results for the six-degree-of-freedom robot optimized for pose goals show the capability of our approach in generating robots for a generic workspace: $97.5\%$ of poses can be reached by the designed robot with an accuracy better than $1$mm and $1\deg$.
While not directly comparable due to differences in setup, our results indicate similar or higher precision than those reported by \citet{Hoffman2025}, but with computation times reduced from hours to only seconds.
This efficiency makes the method practical as a co-pilot, enabling engineers to rapidly iterate over candidate morphologies in a human-in-the-loop workflow.
 
\xhdr{Adapting to Workspace Size}
To analyze the correlation between workspace dimensions and robot size, we uniformly scale workspace dimensions by a factor between $0.5$-$1.5$ in $10$ steps.
We generate 100 robots for each workspace and measure their total size as the sum of Euclidean distances between the proximal and distal module interfaces.
We observe a Pearson correlation coefficient of $0.381$ ($p < 0.01$), indicating a statistically significant positive correlation between workspace size and robot dimensions.

Incorporating goal tolerances into the training, especially when generating underactuated robots, benefits the final performance.
Overall, our approach can generate robot designs to cover a specified workspace with high accuracy while running significantly faster than previous approaches based on iterative optimization.
\textbf{These results undermine hypothesis~\ref{hyp:2}.}
 
\subsection{The Effect of Design Constraints} \label{sec:design_constraints}
\begin{table}
    \small\sf\centering
    \caption{Comparison for the best robot for each design mode across all $36$ tasks 
    \label{tab:mode_comparison}}
    \begin{tabular}{rccc}
    \toprule
         & Solved Tasks & Solved Goals &  Error ($\varnothing$) \\
    \midrule
    Modular & $18$ & $86.56\%$ & $5.2$mm \\
    Economic & $24$ & $92.01\%$ & $1.7$mm \\
    Free & $\bm{28}$ & $\bm{96.53\%}$ & $\bm{1.0}$mm \\
    \bottomrule
    \end{tabular}
\end{table}
We investigate how structural and economic constraints affect robot design optimization by comparing performance across the three design modes introduced in Section~\ref{sec:gen_net}: modular, economical, and free.
Each mode was evaluated in the $36$ tasks introduced in Section~\ref{res:vs_baselines}.
All solutions were collision-free, so our comparison focuses on the goal precision.

As shown in Table~\ref{tab:mode_comparison}, robots designed in free design mode can solve most tasks and goals.
Specifically, $95.53\%$ of goals are solved, representing a substantial performance improvement over the other design modes and the industrial robots analyzed in Section~\ref{res:vs_baselines}, highlighting the advantages of task-tailored robot designs.
Allowing slight adjustments to goal positions of up to $1$cm raises the success rate to $98.3\%$, and permitting at most one goal per task to be moved by $10$cm enables a valid solution for all tasks.
While such adjustments may not always be feasible, these results illustrate the practical potential of our co-pilot in scenarios where human engineers can flexibly adapt the task setup.
Robots designed in economic mode achieved slightly worse results while still being able to solve two-thirds of the tasks and more than $92\%$ of the goals.
As a consequence of the comparatively strong design constraints, the robots generated in the modular mode performed the worst on average, although they still produced valid solutions for half of all tasks.

Analysis of scenarios where less constrained modes outperformed modular designs revealed a common characteristic:
These tasks usually contained at least one target pose positioned close to an obstacle, with orientations that made them difficult or impossible to reach for modular designs.
None of the pre-manufactured robot modules could be placed such that their output reference frame coincides with the goal pose without colliding with a nearby obstacle.
On the other hand, the less constrained free and economical modes could generate robot links capable of reaching these challenging poses without colliding with any obstacle, solving previously impossible reaching tasks.

These results demonstrate that our approach adapts effectively to different design constraints and takes advantage of additional design freedom, achieving the best performance in the least constrained mode. \textbf{Overall, the results support hypothesis~\ref{hyp:3}.}

\subsection{Encoder Ablation}~\label{sec:ablation_encoder}
\begin{figure}
    \centering
    \includegraphics[width=\linewidth]{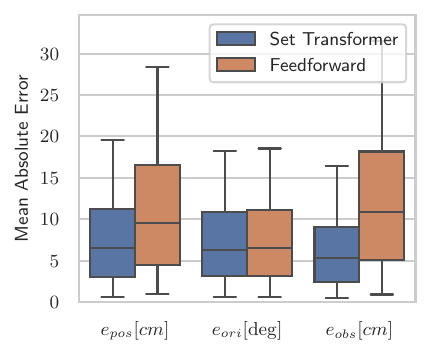}
    \caption{We reconstruct point-set-distances using a set transformer encoder and a feedforward encoder.
    On encodings generated with the set transformer encoder, average errors for both point-set-distances to goal positions and obstacles are lower.
    Boxes represent the $25\%$ and $75\%$ percentiles, whiskers show the $5\%$ and $95\%$ percentiles.
} \label{fig:encoder_ablation-a}
\end{figure}
We investigate the effectiveness of our embedding architecture by replacing the set transformer encoder with a feedforward network followed by a mean pooling operation.
Although this simpler setup remains order-invariant and can deal with an arbitrary number of inputs, the lack of cross-attention mechanisms forbids explicitly encoding relations between different goals and obstacles in the environment.
We train both architectures on the environment reconstruction loss defined in Section~\ref{sec:encoders} exclusively.
To test the expressiveness of the embeddings, we conduct two numerical experiments:
First, we randomly generate $1000$ environments with eight goals and up to $32$ obstacles each.
In addition, we randomly sample $60$ query positions and orientations each.
Then, we reconstruct the set-point distances for the query positions and orientations in the randomly generated environments.
Next, we train a kinematics network to predict collision-free inverse kinematics for a UR20 manipulator, using embeddings from the pre-trained encoder networks.
For training the kinematics network, we generate environments with $32$ obstacles and without any goal, so the kinematics network is trained solely on collision avoidance.
For evaluation, we generate $10000$ of these environments, not all of which might allow a collision-free configuration of the UR20 robot.

Figure~\ref{fig:encoder_ablation-a} shows the mean reconstruction errors for both architectures.
On average, the reconstruction of point-set distances between query and goal positions ($e_{pos}$) improves by $30\%$ when using the set transformer encoder, while the orientation reconstruction error ($e_{ori}$) remains comparable.
Importantly, the set transformer achieves a significantly lower mean absolute error of $6.32$cm when predicting the minimum signed distance between query positions and environment obstacles ($e_{obs}$) -- about half of the $12.66$cm error observed with a linear encoder.

The results of the second experiment highlight the practical relevance of this improvement.
When using embeddings from the linear encoder, the kinematics network is able to produce collision-free configurations for $87.33\%$ of environments.
In contrast, using embeddings from the set transformer, the kinematics network produces $95.06\%$ collision-free configurations.

\subsection{Transfer to the Real World} \label{res:realworld}
\begin{figure}
    \centering
    \includegraphics[width=\linewidth]{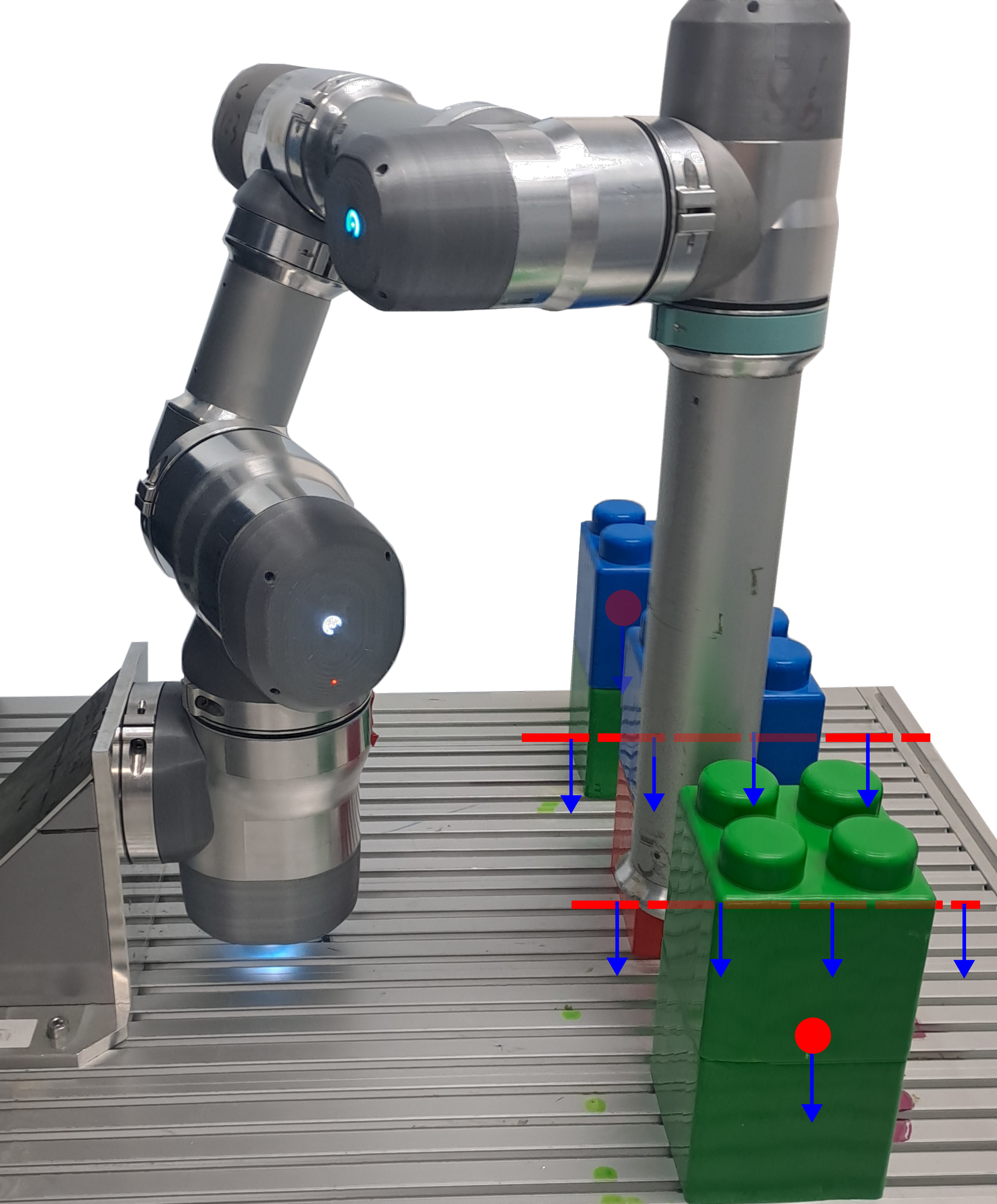}
    \caption{
    A modular robot built from five joint modules and two static extensions:
    A short ``L-shaped'' extension connects the second and the third joint module, while a long ``I-shaped'' extension is mounted after the last joint.
    The dashed red lines and the red spheres represent target positions; blue arrows indicate the target orientation.
    }
    \label{fig:robco_duplo}
\end{figure}
We demonstrate transferability to the real world by setting up a robot generated by a designer network in modular design mode.
The environment for which the robot is optimized is shown in Figure~\ref{fig:robco_duplo}.
It consists of three obstacles and $36$ target poses distributed evenly between them.
We generated five degree-of-freedom designs and selected the first that could be assembled from the limited hardware modules available in our laboratory without further modifications.
The resulting robot is shown in Figure~\ref{fig:robco_duplo} and in the supplementary video.
It could navigate the environment successfully without collisions, achieving an average accuracy of $0.6$mm and a projected orientation error of $3.4\deg$ per target pose, using the tolerance of rotational symmetry.
We took advantage of obtaining up to four different inverse kinematics solutions per pose to obtain a feasible joint-space trajectory.
For every goal, we performed linear interpolation between the chosen joint angles for the previous goal and the closest collision-free IK solution in the joint space for this goal.
Although more sophisticated path-planning algorithms likely would have resulted in a shorter trajectory, this simple heuristic already led to collision-free movement through the obstacle course.

\section{Conclusion}\label{sec:conclusion}
We have introduced a method for generating and optimizing task-tailored manipulators from scratch.
By providing diverse, optimized robot designs almost instantly, our method enables the economic engineering and deployment of task-tailored manipulators.
At its core, we learn a generic IK-solver that supports the co-design of morphology and control.
In this work, we demonstrate its effectiveness on static reachability tasks and a real-world transfer.
By ensuring a differentiable computation of collision and parameter constraints, we train a generative designer network that proposes customized robot morphologies for a specific environment.
In numerical experiments, we demonstrated that the method can produce specialized manipulators across environments, constraints, and goal tolerances.
Our approach achieves accuracy on par with a brute-force approach in many cases and higher than genetic algorithms, but in a fraction of the runtime.
Unlike prior work, it provides optimized designs without requiring extensive ground-truth data or costly iterative search.
Finally, our continuously parameterized designs accelerate the exploration of novel morphologies, and real-world experiments confirm that the method can be directly applied to optimize and control modular robots. By enabling generative optimization over both morphology and task performance, the approach realizes the vision of a design co-pilot for task-tailored manipulators.
Engineers can now interactively explore, adapt, and refine robot morphologies, iterating in response to evolving requirements; as our results show, even minor adjustments to task goals can render previously unreachable tasks solvable, underscoring the practical value of this co-design workflow.

\xhdr{Limitations and future work}
The free and economic design modes can already produce morphologies that are useful for practical deployment, though they may not always be production-ready.
Beyond this, a key advantage is their potential to act as a co-pilot for human engineers, enabling rapid adaptation and iteration of designs to suit specific applications and production constraints.

Our approach relies on numerical optimization to obtain inverse kinematics solutions that are suitable for deployment in industrial applications.
General learning-based IK solvers still struggle to achieve sub-centimeter accuracy and to capture all possible solutions, so the feedback provided to the designer network remains noisy.
Future work could build upon advances in the automatic generation of analytic IK solutions~\citep{Ostermeier2025} to obtain more precise solutions for manipulators designed with our approach.

\section*{Acknowledgments}
The authors gratefully acknowledge financial support by the Horizon 2020 EU Framework Project CONCERT under grant 101016007 and by the Deutsche Forschungsgemeinschaft (German Research Foundation) under grant number AL 1185/31-1.
Further, the authors gratefully acknowledge the scientific support and resources of the AI service infrastructure LRZ AI Systems provided by the Leibniz Supercomputing Centre (LRZ) of the Bavarian Academy of Sciences and Humanities (BAdW), funded by Bayerisches Staatsministerium für Wissenschaft und Kunst (StMWK).
The two engineer icons in Figure~\ref{fig:title_figure} were generated using Microsoft Copilot Chat (web version, September 2025).

\bibliographystyle{plainnat}
\bibliography{bibliography_clean}

\section{Appendix}
\setcounter{figure}{0}
\setcounter{table}{0}
\renewcommand{\thefigure}{A.\arabic{figure}}
\renewcommand{\thetable}{A.\arabic{table}}

\subsection{Collision Pairs}\label{apdx:geometry_approx}
Relevant collision pairs are defined as follows:
Every capsule $c_i$ representing a piece of a manipulator link forms a collision pair with every obstacle in the environment.
In addition, it forms a collision pair with another capsule $c_j$ representing the same robot if not one of the following conditions holds:
\begin{itemize}
    \item $c_i$ and $c_j$ are part of the same rigid body,
    \item $c_i$ and $c_j$ are part of two consecutive rigid bodies in the kinematic chain of the robot, connected by a single joint.
\end{itemize}
The second constraint leads to a slight simplification of the self-collision problem.
However, due to the overapproximation of the robot links, the collision objects of two such bodies always overlap, rendering a differentiable computation of collisions infeasible.
In practice, collision avoidance between consecutive links can often be enforced by setting appropriate joint limits.

\subsection{Link Geometry Parameterisation}\label{apdx:dh_to_links}
\begin{figure}[h!]
    \centering
    \def\svgwidth{\linewidth}
    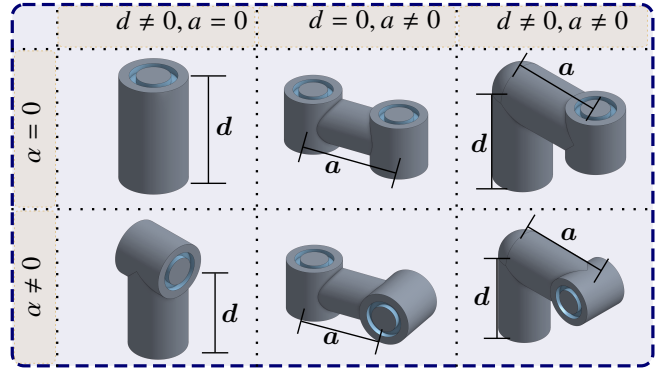
    \caption{Geometric representations of different parameterizations.}
    \label{fig_apdx:dh}
\end{figure}

Figure~\ref{fig_apdx:dh} visualizes the geometries resulting from various parameterizations.
If both $a$ and $d$ are zero, we do not insert a link between two consecutive joints -- this results in a two-dof joint with a spherical body.
The kinematics of the corresponding modules is provided by the standard convention of~\citet{Denavit1955}.

\subsection{Data Representation}\label{apdx:data_representations}
\xhdr{Problem representation}
Each goal is represented by nine values $(x, y, z, r_1, \dots, r_6)$, where the first three are its Cartesian coordinates and $(r_1, \dots, r_6)$ serve as a continuous orientation representation, following~\cite{Zhou2019}.
We overapproximate environmental obstacles using spheres and piecewise cylindrical robot geometries as capsules.
This realizes an efficient and analytic computation of signed distances between the robot and obstacles and a simple four-dimensional obstacle representation consisting of the radius and position of the sphere.
A spherical obstacle is represented by four values $(x, y, z, r)$ that represent the position of its center in Cartesian coordinates, as well as its radius.

\subsection{Hyperparameters}\label{apdx:training_details}
If not specified otherwise, we used the weights and hyperparameters shown in Table~\ref{tab:params}.
For the ablation study presented in Section~\ref{sec:ablation_encoder}, we train both encoders for 100,000 steps.
On a Desktop PC with NVIDIA GeForce RTX 3080 Ti graphics card (12GB memory), Intel i7-11700KF processor (16 cores), we were able to perform between four and twelve steps per second when training the designer network, resulting in training times of 12 to 34 hours for one run.

The baseline genetic algorithm in Section~\ref{res:vs_baselines} is implemented using the EvoTorch library by~\citet{Engin2023}.
We use a single-point crossover, tournament selection with a tournament size of $4$, and random mutation with a per-gene mutation probability of $0.05$.
\begin{table}
\small\sf\centering
\caption{Parameter Values for Numerical Experiments}
\label{tab:params}
\begin{tabular}{ll}
\toprule
\textbf{Parameter} & \textbf{Value} \\
\midrule
Batch size & $128$ \\
Steps (pretraining, IK) & $150,000$ \\
Steps (training) & $350,000$ \\
Learning rate (encoder) & $10^{-5}$ \\
Learning rate (kinematics network) & $10^{-4}$ \\
Learning rate (designer network) & $10^{-4}$ \\
Numerical IK, maximum steps & $50$ \\
Optimization Steps (Adam) & $200$ \\
Learning rate (optimization, IK) & $0.1$ \\
Learning rate (optimization, parameters) & $0.01$ \\
\midrule
(Self-) attention heads (designer) & $8$ each \\
(Block-) attention heads (encoders) & $8$ each \\
Designer decoder layers & $8$ \\
Set transformer encoder layers & $4$ \\
\midrule
Weight $w_d$ (pose distance) [m] & $5$ \\
Weight $w_r$ (rotational distance) [\(\deg\)] & $0.5$ \\
Weight $w_{col}$ (collision loss) & $0.6$ \\
Weight $w_{IK}$ (cosine similarity IK heads) & $0.4$ \\
Weight $w_g$ (robot similarity) & $0.5$ \\
Weight $w_e$ (cross-environment similarity) & $2.0$ \\
Weight $w_r$ (regularization) & $1.0$ \\
Weight $w_c$ (hardware cost) & $0.1$ \\
\midrule
Threshold (collision loss) [mm] & $120$ \\
Hardware Cost $c_r$& $\left( \sum a_i + \sum d_i \right)$ \\
Diameter of robot links [mm] & $120$ \\
Gumbel scheduling & Linear \\
Gumbel temperature start & $3$ \\
Gumbel temperature end & $0.01$ \\
\bottomrule
\end{tabular}
\end{table}

\subsection{Details on Brute-Force Approach Results}\label{apdx:bf_details}
\begin{figure*}
    \centering
    \includegraphics[width=0.9\linewidth]{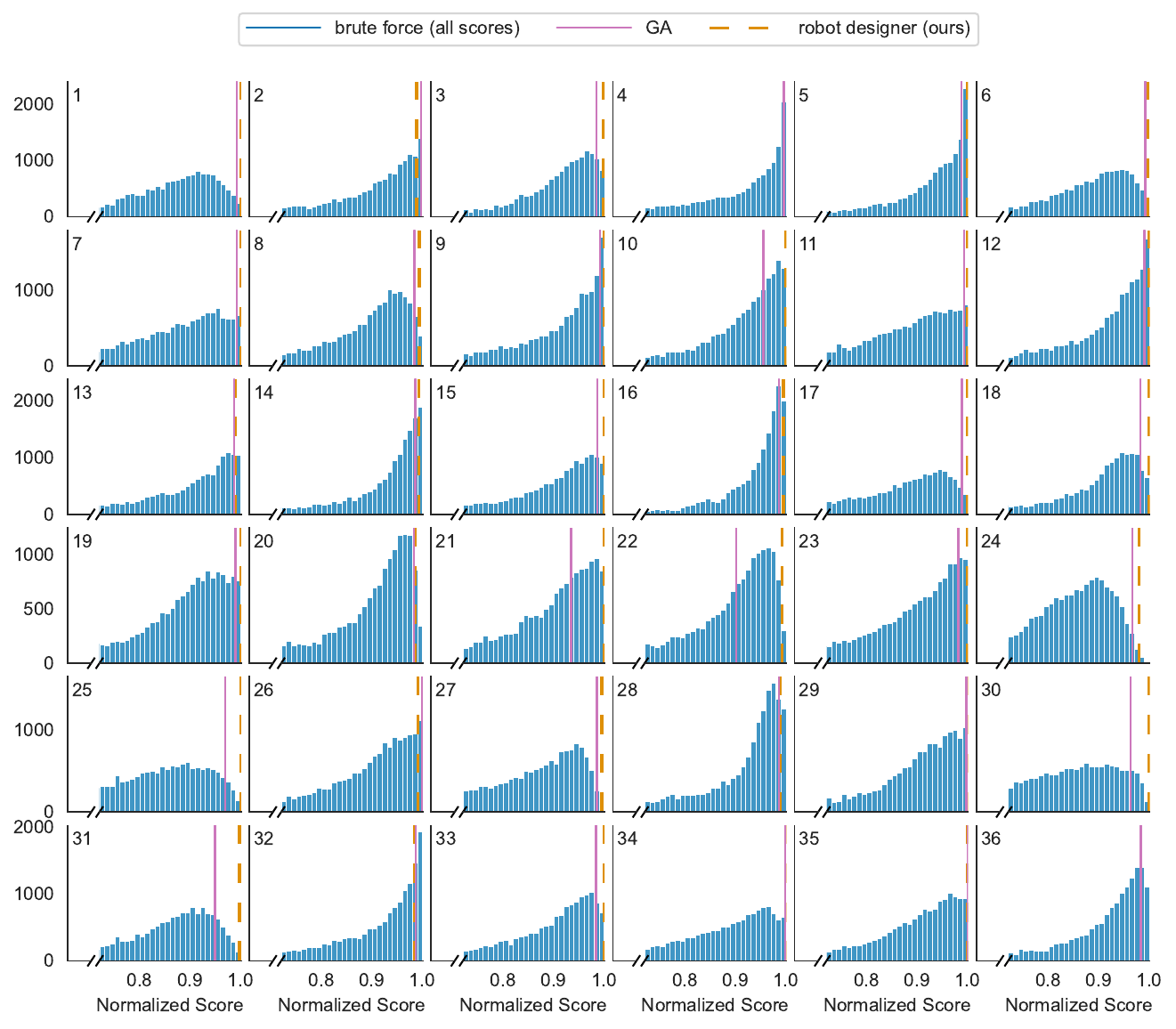}
    \caption{
    Normalized scores for the brute-force approach and the best robot returned by our approach across all $36$ tasks.
    In $31$ tasks, our approach outperforms the genetic algorithm despite requiring a fraction of the runtime.
    }
    \label{fig:bf_cluttered_comparison}
\end{figure*}

Figure~\ref{fig:bf_cluttered_comparison} shows a histogram of normalized fitness scores for each of the $36$ environments introduced in Section~\ref{res:vs_baselines} individually.

\subsection{Comparison of Experimental Setups}\label{apdx:experiment_comparison}
In Section~\ref{res:goal_tolerances}, we consider a setup similar to that presented by~\citet{Hoffman2025}.
However, there are some differences that must be taken into account when comparing the results, listed below.
\begin{itemize}
    \item There is no publicly available data for the Tiago robot used by Hoffman et al., so our work reproduces their experiment using different robot modules. We scaled the original workspace by a factor of $0.8$ to account for the difference in the size and shape of the robot modules.
    \item The work by Hoffman et al. considers static torque constraints while our method focuses on robot kinematics.
    \item Hoffman et al. only consider the reachability of Euclidean workspace positions, while our method generalizes to $SE(3)$ poses.
    \item In contrast to Hoffman et al., we consider the self-collisions of the robot. In addition, we show that our method can be generalized to workspaces with obstacles.
    \item Our method initially requires a computationally expensive training phase, whereas inference for a specific problem takes only a few seconds. In contrast, the method presented by Hoffman et al. does not require a training phase. However, they report run-times of multiple hours to generate a solution for a new problem.
\end{itemize}

\end{document}